\documentclass[11pt,a4paper]{article}
\usepackage[hyperref]{acl2019}
\usepackage{times}
\usepackage{latexsym}

\usepackage{graphicx}
\usepackage{url}
\usepackage{bm}
\usepackage{amssymb,amsfonts,amsmath}
\usepackage{algorithm}
\usepackage{algorithmicx}
\usepackage{algpseudocode}
\usepackage{mathtools}
\usepackage{booktabs}
\usepackage{subcaption}
\usepackage{xspace}
\usepackage{dcolumn}
\usepackage{comment}

\newcommand{\cev}[1]{\reflectbox{\ensuremath{\vec{\reflectbox{\ensuremath{#1}}}}}}

\DeclareMathOperator*{\argmax}{arg\,max}

\newcolumntype{d}[1]{D{.}{.}{#1}}

\aclfinalcopy 
\def\aclpaperid{1341} 

\newcommand\Pivot{P\textsc{ivot}\xspace}

\title{Key Fact as Pivot: A Two-Stage Model for Low Resource \\Table-to-Text Generation}

\author{
Shuming Ma,\textsuperscript{\rm 1,3}
Pengcheng Yang,\textsuperscript{\rm 1,2}
Tianyu Liu,\textsuperscript{\rm 1}
Peng Li,\textsuperscript{\rm 3}
Jie Zhou,\textsuperscript{\rm 3}
Xu Sun\textsuperscript{\rm 1,2}\\
\textsuperscript{\rm 1}MOE Key Lab of Computational Linguistics, School of EECS, Peking University\\
\textsuperscript{\rm 2}Deep Learning Lab, Beijing Institute of Big Data Research, Peking University\\
\textsuperscript{\rm 3}Pattern Recognition Center, WeChat AI, Tencent Inc, China\\
\{shumingma,yang\_pc,tianyu0421,xusun\}@pku.edu.cn \\
\{patrickpli,withtomzhou\}@tencent.com
}

\date{}

\begin{document}
\maketitle
\begin{abstract}
  Table-to-text generation aims to translate the structured data into the unstructured text. Most existing methods adopt the encoder-decoder framework to learn the transformation, which requires large-scale training samples. However, the lack of large parallel data is a major practical problem for many domains. In this work, we consider the scenario of low resource table-to-text generation, where only limited parallel data is available. We propose a novel model to separate the generation into two stages: key fact prediction and surface realization. It first predicts the key facts from the tables, and then generates the text with the key facts. The training of key fact prediction needs much fewer annotated data, while surface realization can be trained with pseudo parallel corpus. We evaluate our model on a biography generation dataset. Our model can achieve $27.34$ BLEU score with only $1,000$ parallel data, while the baseline model only obtain the performance of $9.71$ BLEU score.\footnote{The codes are available at \url{https://github.com/lancopku/Pivot}.} 
\end{abstract}

\section{Introduction}

\begin{figure}[t]
	\centering
	\includegraphics[width=1.0\linewidth]{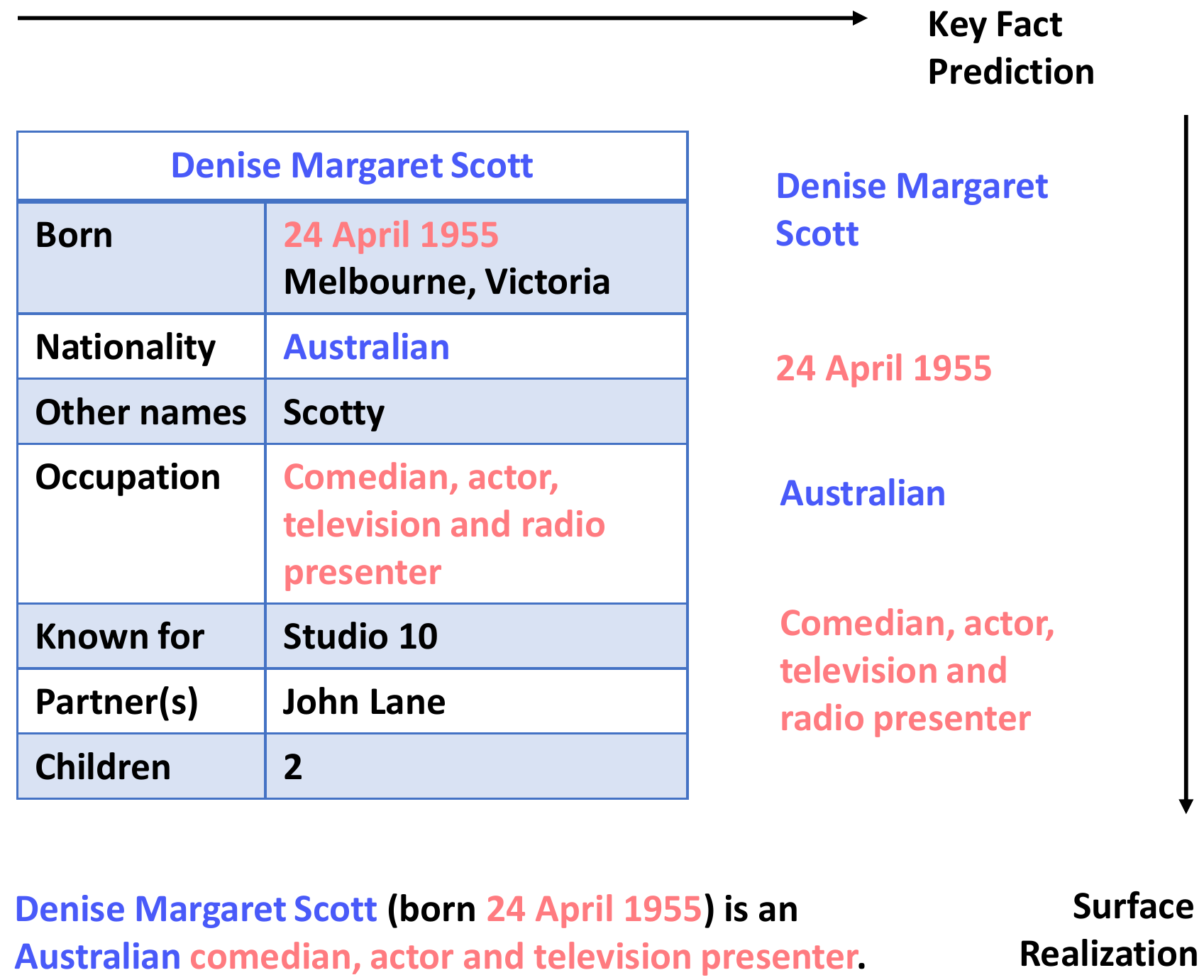}
	\caption{An example of table-to-text generation, and also a flow chart of our method.}\label{fig:example}
\end{figure}

Table-to-text generation is to generate a description from the structured table. It helps readers to summarize the key points in the table, and tell in the natural language. Figure~\ref{fig:example} shows an example of table-to-text generation. The table provides some structured information about a person named ``Denise Margaret Scott'', and the corresponding text describes the person with the key information in the table. Table-to-text generation can be applied in many scenarios, including weather report generation~\cite{Liang2009Learning}, NBA news writing~\cite{Barzilay2005Collective}, biography generation~\cite{Dubou2002Content,Lebret2016Neural}, and so on. Moreover, table-to-text generation is a good testbed of a model's ability of understanding the structured knowledge.

\begin{figure}[t]
	\centering
	\includegraphics[width=1.0\linewidth]{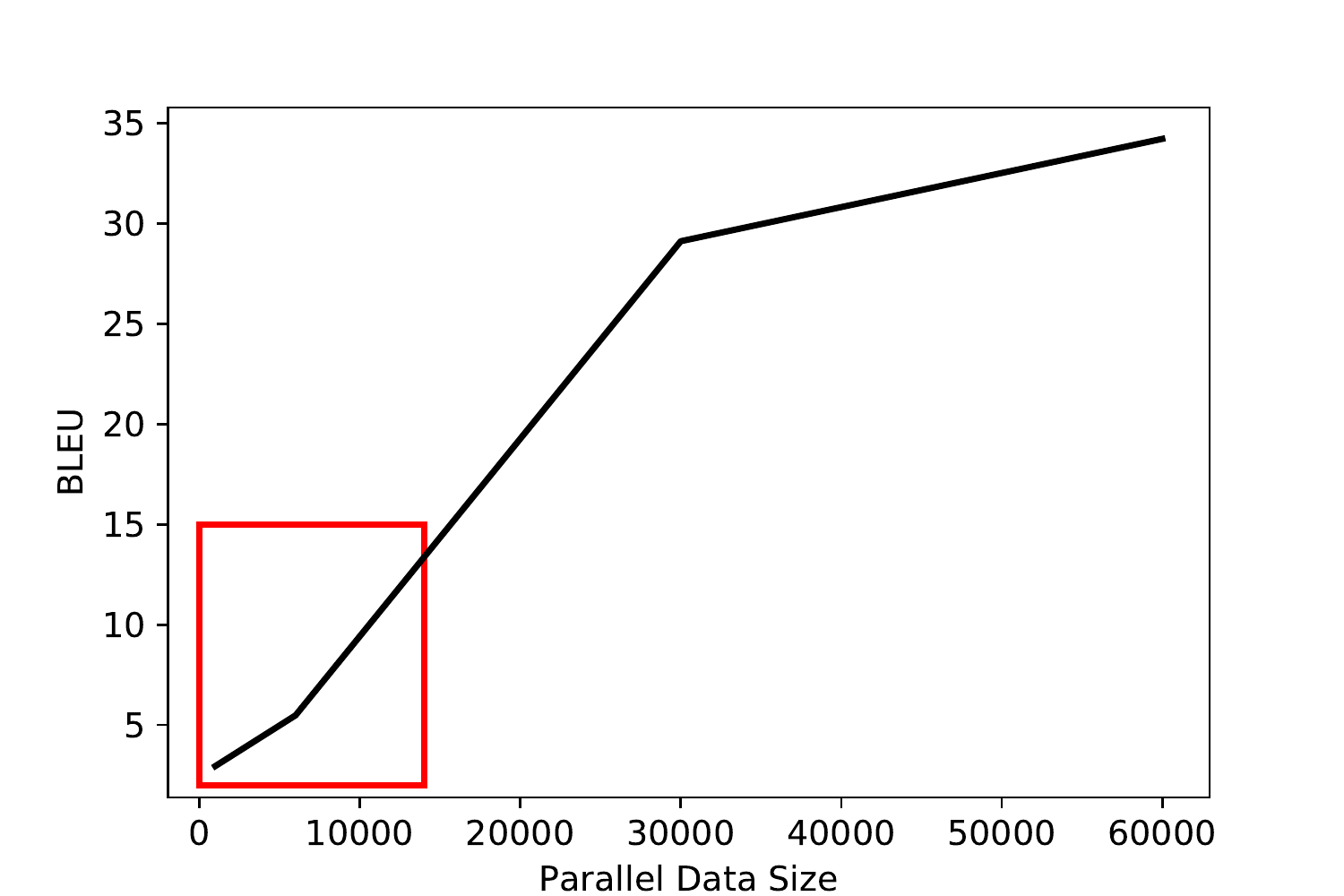}
	\caption{The BLEU scores of the a table-to-text model trained with different number of parallel data under the encoder-decoder framework on the WIKIBIO dataset.}\label{fig:vanilla}
\end{figure}

Most of the existing methods for table-to-text generation are based on the encoder-decoder framework~\cite{seq2seq,attention}. They represent the source tables with a neural encoder, and generate the text word-by-word with a decoder conditioned on the source table representation. Although the encoder-decoder framework has proven successful in the area of natural language generation (NLG)~\cite{stanfordattention,ras,lu17,yang2018sgm}, it requires a large parallel corpus, and is known to fail when the corpus is not big enough. Figure~\ref{fig:vanilla} shows the performance of a table-to-text model trained with different number of parallel data under the encoder-decoder framework. We can see that the performance is poor when the parallel data size is low. In practice, we lack the large parallel data in many domains, and it is expensive to construct a high-quality parallel corpus.


This work focuses on the task of low resource table-to-text generation, where only limited parallel data is available. Some previous work~\citep{Puduppully2018D2T,Gehrmann2018E2E} formulates the task as the combination of content selection and surface realization, and models them with an end-to-end model. Inspired by these work, we break up the table-to-text generation into two stages, each of which is performed by a model trainable with only a few annotated data. Specifically, it first predicts the key facts from the tables, and then generates the text with the key facts, as shown in Figure~\ref{fig:example}. 
The two-stage method consists of two separate models: a key fact prediction model and a surface realization model. The key fact prediction model is formulated as a sequence labeling problem, so it needs much fewer annotated data than the encoder-decoder models. According to our experiments, the model can obtain $87.92\%$ F1 score with only $1,000$ annotated data. As for the surface realization model, we propose a method to construct a pseudo parallel dataset without the need of labeled data. In this way, our model can make full use of the unlabeled text, and alleviate the heavy need of the parallel data.

The contributions of this work are as follows:
\begin{itemize}
    \item We propose to break up the table-to-text generation into two stages with two separate models, so that the model can be trained with fewer annotated data. 
    \item We propose a method to construct a pseudo parallel dataset for the surface realization model, without the need of labeled data.
    \item Experiments show that our proposed model can achieve $27.34$ BLEU score on a biography generation dataset with only $1,000$ table-text samples.
\end{itemize}

\begin{figure*}[t]
	\centering
	\includegraphics[width=1.0\linewidth]{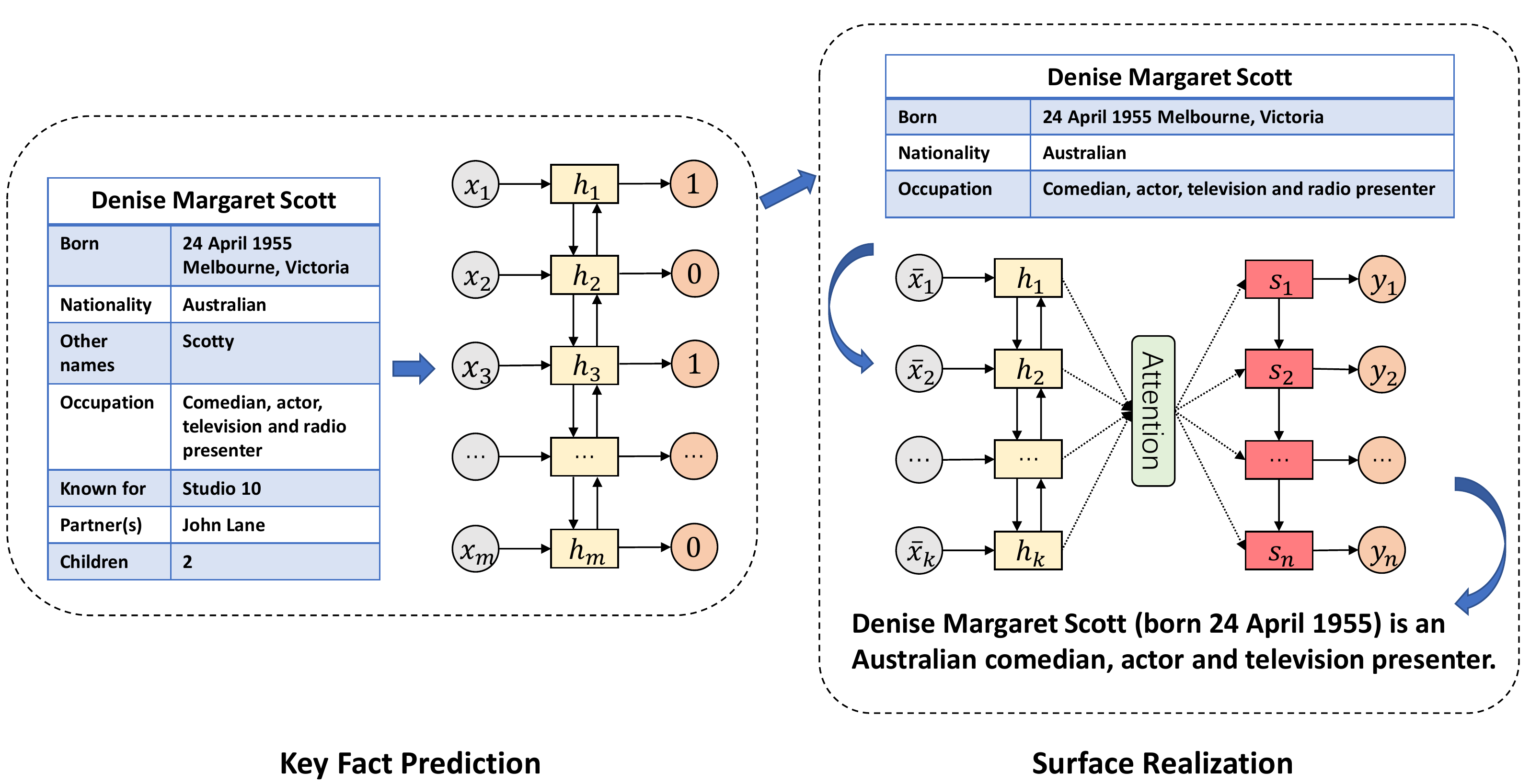}
	\caption{The overview of our model. For illustration, the surface realization model is a vanilla Seq2Seq, while it can also be a Transformer in our implementation.}\label{fig:overview}
\end{figure*}

\section{\Pivot: A Two-Stage Model}

In this section, we introduce our proposed two-stage model, which we denote as \textbf{\Pivot}. We first give the formulation of the table-to-text generation and the related notations. Then, we provide an overview of the model. Finally, we describe the two models for each stage in detail.

\subsection{Formulation and Notations}


Suppose we have a parallel table-to-text dataset $\mathcal{P}$ with $N$ data samples and an unlabeled text dataset $\mathcal{U}$ with $M$ samples. Each parallel sample consists of a source table $T$ and a text description $\bm{y} = \{y_1, y_2, \cdots, y_n\}$. 
The table $T$ can be formulated as $K$ records $T = \{r_1, r_2, r_3, \cdots, r_K\}$, and each record is an attribute-value pair $r_j = (a_j, v_j)$. Each sample in the unlabeled text dataset $\mathcal{U}$ is a piece of text $\bm{\bar{y}} = \{\bar{y}_1, \bar{y}_2, \cdots, \bar{y}_n\}$.

Formally, the task of table-to-text generation is to take the structured representations of table $T = \{(a_1, v_1), (a_2, v_2), \cdots, (a_m, v_m)\}$ as input, and output the sequence of words $\bm{y} = \{y_1, y_2, \cdots, y_n\}$. 

\subsection{Overview}

Figure~\ref{fig:overview} shows the overview architecture of our proposed model. Our model contains two stages: key fact prediction and surface realization. At the first stage, we represent the table into a sequence, and use a table-to-pivot model to select the key facts from the sequence. The table-to-pivot model adpots a bi-directional Long Short-term Memory Network (Bi-LSTM) to predict a binary sequence of whether each word is reserved as the key facts. At the second stage, we build a sequence-to-sequence model to take the key facts selected in the first stage as input and emit the table description. In order to make use of the unlabeled text corpus, we propose a method to construct pseudo parallel data to train a better surface realization model. Moreover, we introduce a denoising data augmentation method to reduce the risk of error propagation between two stages.

\subsection{Preprocessing: Key Fact Selection}

The two stages are trained separately, but we do not have the labels of which words in the table are the key facts in the dataset. In this work, we define the co-occurrence facts between the table and the text as the key facts, so we can label the key facts automatically. Algorithm~\ref{alg:pivot} illustrates the process of automatically annotating the key facts. Given a table and its associated text, we enumerate each attribute-value pair in the table, and compute the word overlap between the value and the text. The word overlap is defined as the number of words that are not stop words or punctuation but appear in both the table and the text. We collect all values that have at least one overlap with the text, and regard them as the key facts. In this way, we can obtain a binary sequence with the $0/1$ label denoting whether the values in the table are the key facts. The binary sequence will be regarded as the supervised signal of the key fact prediction model, and the selected key facts will be the input of the surface realization model.

\begin{algorithm}[t] 
	\small
	\caption{Automatic Key Fact Annotation}\label{alg:pivot}
	\begin{algorithmic}[1]
	    \Require A parallel corpora $\mathcal{P}=\{(\bm{x}_i, \bm{y}_i)\}$, where $\bm{x}_i$ is a table, and $\bm{y}_i$ is a word sequence.
	    \State Initial the selected key fact list $\mathbb{W}=[]$
	    
	\For {each sample $(\bm{x}, \bm{y})$ in the parallel dataset $\mathcal{P}$}
		\State $\bm{x} = \{(v_1, a_1), (v_2, a_2), \cdots, (v_m, a_m)\}$
		\State $\bm{y} = \{y_1, y_2, \cdots, y_n\}$
	    \State Initial the selected attribute set $\mathbb{A}=\{\}$
	    \State Initial the selected key fact list $\mathbb{W}_i=[]$
		\For {each attribute-value pair $(v_i, a_i)$ in table $\bm{x}$}
		    \If{$v_i$ in $\bm{y}$ \textbf{And} $v_i$ is not stop word}
		        \State Append attribute $a_i$ into attribute set $\mathbb{A}$
		    \EndIf
		    \If{$a_i$ in $\mathbb{A}$}
		        \State Append value $v_i$ into key fact list $\mathbb{W}_i$
		    \EndIf
		\EndFor
		\State Collect the key fact list $\mathbb{W}$ += $\mathbb{W}_i$
	\EndFor
	
	    \Ensure The selected key fact list $\mathbb{W}$
	\end{algorithmic} 
\end{algorithm} 

\subsection{Stage 1: Key Fact Prediction}

The key fact prediction model is a Bi-LSTM layer with a multi-layer perceptron (MLP) classifier to determine whether each word is selected.
In order to represent the table, we follow the previous work~\cite{Liu2018T2T} to concatenate all the words in the values of the table into a word sequence, and each word is labeled with its attribute. In this way, the table is represented as two sequences: the value sequence $\{v_1, v_2, \cdots, v_m\}$ and the attribute sequence $\{a_1, a_2, \cdots, a_m\}$. A word embedding and an attribute embedding are used to transform two sequences into the vectors. 
Following~\cite{Lebret2016Neural,Liu2018T2T}, we introduce a position embedding to capture structured information of the table.
The position information is represented as a tuple ($p^{+}_{w}$, $p^{−}_{w}$), which includes the positions of the token $w$ counted from the beginning and the end of the value respectively. 
For example, the record of ``(\emph{Name, Denise Margaret Scott})'' is represented as ``(\emph{\{Denise, Name, 1, 3\}, \{Margaret, Name, 2, 2\}, \{Scott, Name, 3, 1\}})''. In this way, each token in the table has an unique feature embedding even if there exists two same words.
Finally, the word embedding, the attribute embedding, and the position embedding are concatenated as the input of the model $\bm{x}$.

\noindent \textbf{Table Encoder:} The goal of the source table encoder is to provide a series of representations for the classifier. More specifically, the table encoder is a Bi-LSTM:
\begin{align}\label{lstm}
h_t = \mathrm{BiLSTM}(x_t,\vec{h}_{t-1},\cev{h}_{t+1})
\end{align}
where $\vec{h}_t$ and $\cev{h}_t$ are the forward and the backward hidden outputs respectively,  $h_t$ is the concatenation of $\vec{h}_t$ and $\cev{h}_t$, and $x_t$ is the input at the $t$-th time step.

\noindent \textbf{Classifier:} The output vector $h_t$ is fed into a MLP classifier to compute the probability distribution of the label $p_1(l_t|\bm{x})$
\begin{align}\label{classifier}
p_1(l_t|\bm{x})=\mathrm{softmax}{(W_{c}h_t+b_{c})}
\end{align}
where $W_{c}$ and $b_{c}$ are trainable parameters of the classifier.

\subsection{Stage 2: Surface Realization}

The surface realization stage aims to generate the text conditioned on the key facts predicted in Stage 1. We adpot two models as the implementation of surface realization: the vanilla Seq2Seq and the Transformer~\cite{Vaswani2017transformer}.

\paragraph{Vanilla Seq2Seq:} In our implementation, the vanilla Seq2Seq consists of a Bi-LSTM encoder and an LSTM decoder with the attention mechanism. The Bi-LSTM encoder is the same as that of the key fact prediction model, except that it does not use any attribute embedding or position embedding. 

The decoder consists of an LSTM, an attention component, and a word generator. It first generates the hidden state $s_t$:
\begin{align}\label{flstm}
s_t = f(y_{t-1},s_{t-1})
\end{align}
where $f(\cdot,\cdot)$ is the function of LSTM for one time step, and $y_{t-1}$ is the last generated word at time step $t-1$.
Then, the hidden state $s_t$ from LSTM is fed into the attention component:
\begin{align}\label{attention}
v_{t}=\mathrm{Attention}({s_t, \bm{h}})
\end{align}
where $\mathrm{Attention}(\cdot,\cdot)$ is the implementation of global attention in \citep{stanfordattention}, and $\bm{h}$ is a sequence of outputs by the encoder.

Given the output vector $v_{t}$ from the attention component, the word generator is used to compute the probability distribution of the output words at time step $t$:
\begin{align}\label{generator}
p_2(y_t|x)=\mathrm{softmax}{(W_{g}v_{t}+b_{g})}
\end{align}
where $W_{g}$ and $b_{g}$ are parameters of the generator. The word with the highest probability is emitted as the $t$-th word.

\paragraph{Transformer:}

Similar to vanilla Seq2Seq, the Transformer consists of an encoder and a decoder. The encoder applies a Transformer layer to encode each word into the representation $h_t$:
\begin{align}
h_t = \mathrm{Transformer}(x_t, \bm{x})
\end{align}
Inside the Transformer, the representation $x_t$ attends to a collection of the other representations $\bm{x}=\{x_1,x_2,\cdots,x_m\}$.
Then, the decoder produces the hidden state by attending to both the encoder outputs and the previous decoder outputs:
\begin{align}
v_t = \mathrm{Transformer}(y_{t}, \bm{y_{<t}}, \bm{h})
\end{align}
Finally, the output vector is fed into a word generator with a softmax layer, which is the same as Eq.~\ref{generator}.

For the purpose of simplicity, we omit the details of the inner computation of the Transformer layer, and refer the readers to the related work~\cite{Vaswani2017transformer}.

\subsection{Pseudo Parallel Data Construction}
\label{sec:pseudo}

The surface realization model is based on the encoder-decoder framework, which requires a large amount of training data. In order to augment the training data, we propose a novel method to construct  pseudo parallel data. The surface realization model is used to organize and complete the text given the key facts. Therefore, it is possible to construct the pseudo parallel data by removing the skeleton of the text and reserving only the key facts. In implementation, we label the text with Stanford CoreNLP toolkit\footnote{\url{https://stanfordnlp.github.io/CoreNLP/index.html}} to assign the POS tag for each word. We reserve the words whose POS tags are among the tag set of \{\emph{NN, NNS, NNP, NNPS, JJ, JJR, JJS, CD, FW}\}, and remove the remaining words. In this way, we can construct a large-scale pseudo parallel data to train the surface realization model.

\subsection{Denoising Data Augmentation}

A problem of the two-stage model is that the error may propagate from the first stage to the second stage. A possible solution is to apply beam search to enlarge the searching space at the first stage. However, in our preliminary experiments, when the beam size is small, the diversity of predicted key facts is low, and also does not help to improve the accuracy. When the beam size is big, the decoding speed is slow but the improvement of accuracy is limited.

To address this issue, we implement a method of denoising data augmentation to reduce the hurt from error propagation and improve the robustness of our model. In practice, we randomly drop some words from the input of surface realization model, or insert some words from other samples. The dropping simulates the cases when the key fact prediction model fails to recall some co-occurrence, while the inserting simulates the cases when the model predicts some extra facts from the table. By adding the noise, we can regard these data as the adversarial examples, which is able to improve the robustness of the surface realization model.

\subsection{Training and Decoding}

Since the two components of our model are separate, the objective functions of the models are optimized individually. 

\paragraph{Training of Key Fact Prediction Model:} 
The key fact prediction model, as a sequence labeling model, is trained using the cross entropy loss:
\begin{align}\label{eq:loss1}
L_{1} = -\sum_{i=1}^{m}\log p_1(l_i|\bm{x})
\end{align}

\paragraph{Training of Surface Realization Model:}
The loss function of the surface realization model can be written as:
\begin{align}\label{eq:loss2}
L_{2} = -\sum_{i=1}^{n}\log p_2(y_i|\bm{\bar{x}})
\end{align}
where $\bm{\bar{x}}$ is a sequence of the selected key facts at Stage 1.
The surface realization model is also trained with the pseudo parallel data as described in Section~\ref{sec:pseudo}. The objective function can be written as:
\begin{align}\label{eq:loss3}
L_{3} = -\sum_{i=1}^{n}\log p_2(\bar{y}_i|\bm{\hat{x}})
\end{align}
where $\bar{y}$ is the unlabeled text, and $\bm{\hat{x}}$ is the pseudo text paired with $\bar{y}$.

\paragraph{Decoding:} 
The decoding consists of two steps. At the first step, it predicts the label by the key fact prediction model:
\begin{align}
\hat{l}_t = \argmax_{l_t \in \{0,1\}}{p_1(l_t|\bm{x})}
\end{align}
The word with $\hat{l}_t=1$ is reserved, while that with $\hat{l}_t=0$ is discarded. Therefore, we can obtain a sub-sequence $\bm{\bar{x}}$ after the discarding operation.

At the second step, the model emits the text with the surface realization model:
\begin{align}
\hat{y}_t = \argmax_{y_t \in \mathcal{V}}p_2(y_t|\bm{\bar{x}})
\end{align}
where $\mathcal{V}$ is the vocabulary size of the model. Therefore, the word sequence $\{\hat{y}_1, \hat{y}_2, \cdots, \hat{y}_N\}$ forms the generated text.

\section{Experiments}

We evaluate our model on a table-to-text generation benchmark. We denote the \Pivot model under the vanilla Seq2Seq framework as \Pivot-Vanilla, and that under the Transformer framework as \Pivot-Trans.

\subsection{Dataset}

We use WIKIBIO dataset~\citep{Lebret2016Neural} as our benchmark dataset. The dataset contains $728,321$ articles from English Wikipedia, which uses the first sentence of each article as the description of the related infobox. There are an average of $26.1$ words in each description, of which $9.5$ words also appear in the table. The table contains $53.1$ words and $19.7$ attributes on average. Following the previous work~\citep{Lebret2016Neural, Liu2018T2T}, we split the dataset into $80\%$ training set, $10\%$ testing set, and $10\%$ validation set. In order to simulate the low resource scenario, we randomly sample $1,000$ parallel sample, and remove the tables from the rest of the training data.

\subsection{Evaluation Metrics}

Following the previous work~\citep{Lebret2016Neural,Wiseman2018Template}, we use BLEU-4~\citep{bleu}, ROUGE-4 (F measure)~\citep{rouge}, and NIST-4~\citep{nist} as the evaluation metrics.

\subsection{Implementation Details}

The vocabulary is limited to the $20,000$ most common words in the training dataset. The batch size is $64$ for all models. We implement the early stopping mechanism with a patience that the performance on the validation set does not fall in $4$ epochs. We tune the hyper-parameters based on the performance on the validation set. 

The key fact prediction model is a Bi-LSTM. The dimensions of the hidden units, the word embedding, the attribute embedding, and the position embedding are $500$, $400$, $50$, and $5$, respectively.

We implement two models as the surface realization models. For the vanilla Seq2Seq model, we set the hidden dimension, the embedding dimension, and the dropout rate~\citep{dropout} to be $500$, $400$, and $0.2$, respectively. For the Transfomer model, the hidden units of the multi-head component and the feed-forward layer are $512$ and $2048$. The embedding size is $512$, the number of heads is $8$, and the number of Transformer blocks is $6$. 

We use the Adam~\cite{kingma2014adam} optimizer to train the models. For the hyper-parameters of Adam optimizer, we set the learning rate $\alpha = 0.001$, two momentum parameters $\beta_{1} = 0.9$ and $\beta_{2} = 0.999$, and $\epsilon = 1 \times 10^{-8}$. We clip the gradients~\cite{gradientclip} to the maximum norm of $5.0$. We half the learning rate when the performance on the validation set does not improve in $3$ epochs.

\subsection{Baselines}

We compare our models with two categories of baseline models: the supervised models which exploit only parallel data (Vanilla Seq2Seq, Transformer, Struct-aware), and the semi-supervised models which are trained on both parallel data and unlabelled data (PretrainedMT, SemiMT). The baselines are as follows:
\begin{itemize}
\item  \textbf{Vanilla Seq2Seq}~\citep{seq2seq} with the attention mechanism~\citep{attention} is a popular model for natural language generation. 

\item \textbf{Transformer}~\citep{Vaswani2017transformer} is a state-of-the-art model under the encoder-decoder framework, based solely on attention mechanisms.

\item \textbf{Struct-aware}~\citep{Liu2018T2T} is the state-of-the-art model for table-to-text generation. It models the inner structure of table with a field-gating mechanism insides the LSTM, and learns the interaction between tables and text with a dual attention mechanism.

\item \textbf{PretrainedMT}~\citep{Skorokhodov2018Semi} is a semi-supervised method to pretrain the decoder of the sequence-to-sequence model with a language model.

\item \textbf{SemiMT}~\citep{Cheng2016Semi} is a semi-supervised method to jointly train the sequence-to-sequence model with an auto-encoder.

\end{itemize}

The supervised models are trained with the same parallel data as our model, while the semi-supervised models share the same parallel data and the unlabeled data as ours.

\begin{table}[t]
\centering
\small
\begin{tabular}{l d{2.2} c d{2.2}}
\toprule
\multicolumn{1}{c}{\bf Model} & \multicolumn{1}{c}{\bf F1} & \bf P & \multicolumn{1}{c}{\bf R} \\ 
\midrule
\Pivot (Bi-LSTM) & 87.92 & 92.59 & 83.70  \\
\bottomrule
\\
\toprule
\multicolumn{1}{c}{\bf Model} & \multicolumn{1}{c}{\bf BLEU} & \bf NIST & \multicolumn{1}{c}{\bf ROUGE} \\ 
\midrule
Vanilla Seq2Seq & 2.14 & 0.2809 & 0.47 \\
Structure-S2S & 3.27 & 0.9612 & 0.71 \\
\midrule
PretrainedMT &  4.35 & 1.9937 & 0.91 \\
SemiMT & 6.76  & 3.5017 & 2.04 \\
\midrule
\Pivot-Vanilla & 20.09 & 6.5130 & 18.31 \\
\bottomrule
\\
\toprule
\multicolumn{1}{c}{\bf Model} & \multicolumn{1}{c}{\bf BLEU} & \bf NIST & \multicolumn{1}{c}{\bf ROUGE} \\ 
\midrule
Transformer & 5.48 & 1.9873 & 1.26 \\
\midrule
PretrainedMT & 6.43 & 2.1019 & 1.77 \\
SemiMT & 9.71 & 2.7019 & 3.31 \\
\midrule
\Pivot-Trans & 27.34 & 6.8763 & 19.30  \\
\bottomrule
\end{tabular}
\caption{Results of our model and the baselines. Above is the performance of the key fact prediction component (F1: F1 score, P: precision, R: recall). Middle is the comparison between models under the Vanilla Seq2Seq framework. Below is the models implemented with the transformer framework.}
\label{table:result}

\end{table}

\subsection{Results}

We compare our \Pivot model with the above baseline models. Table~\ref{table:result} summarizes the results of these models. It shows that our \Pivot model achieves $87.92\%$ F1 score, $92.59\%$ precision, and $83.70\%$ recall at the stage of key fact prediction, which provides a good foundation for the stage of surface realization. Based on the selected key facts, our models achieve the scores of $20.09$ BLEU, $6.5130$ NIST, and $18.31$ ROUGE under the vanilla Seq2Seq framework, and $27.34$ BLEU, $6.8763$ NIST, and $19.30$ ROUGE under the Transformer framework, which significantly outperform all the baseline models in terms of all metrics. Furthermore, it shows that the implementation with the Transformer can obtain higher scores than that with the vanilla Seq2Seq.

\begin{figure}[t]
	\centering
	\subcaptionbox{Vanilla Seq2Seq v.s \Pivot-Vanilla}{\includegraphics[width=1.0\linewidth]{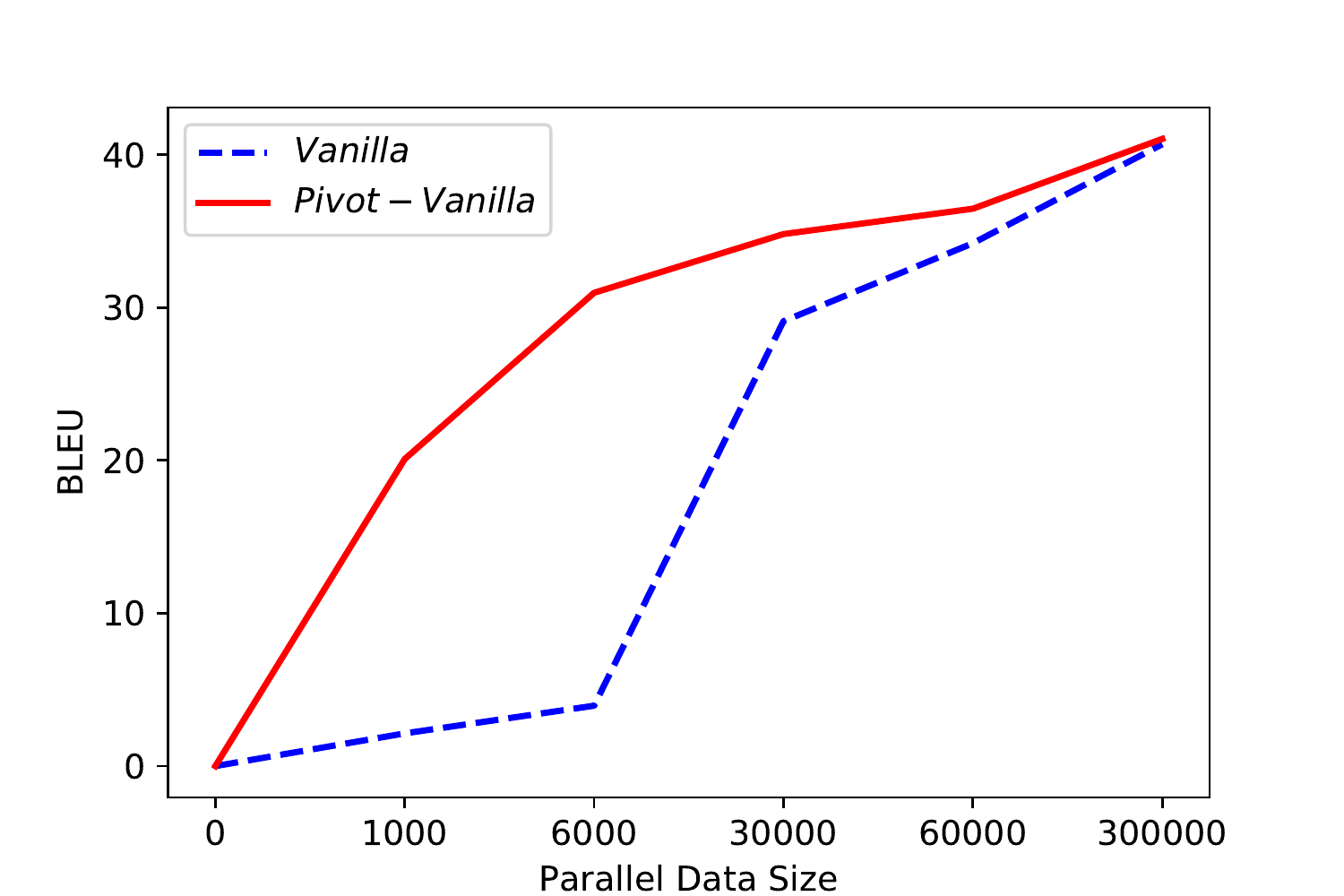}}
	\subcaptionbox{Transformer v.s \Pivot-Trans}{\includegraphics[width=1.0\linewidth]{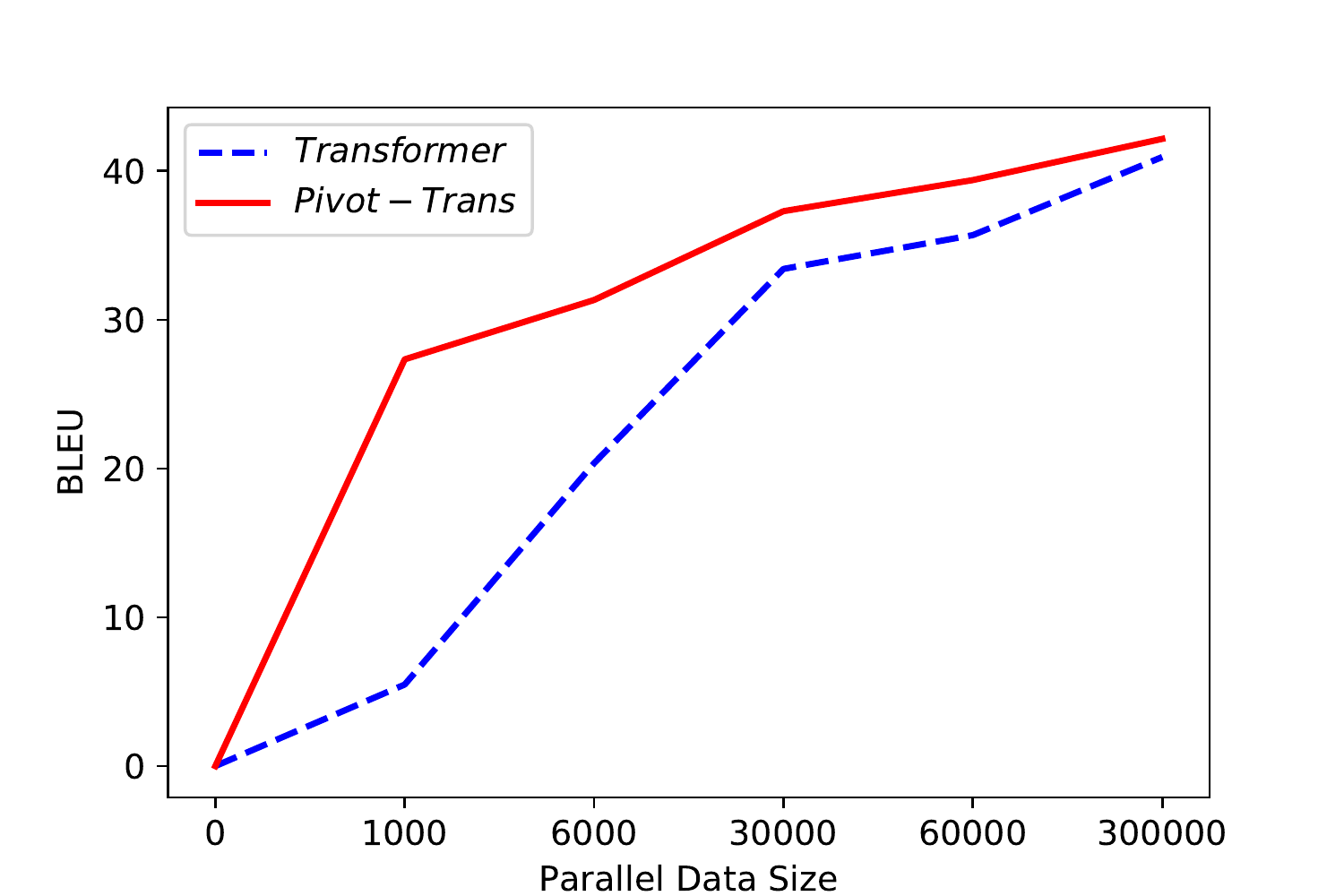}}
	\caption{The BLEU measure of our Pivot model and the baselines trained with different parallel data size.}\label{fig:pivot_curve}
\end{figure}

\begin{figure}[t]
	\centering
	\includegraphics[width=1.0\linewidth]{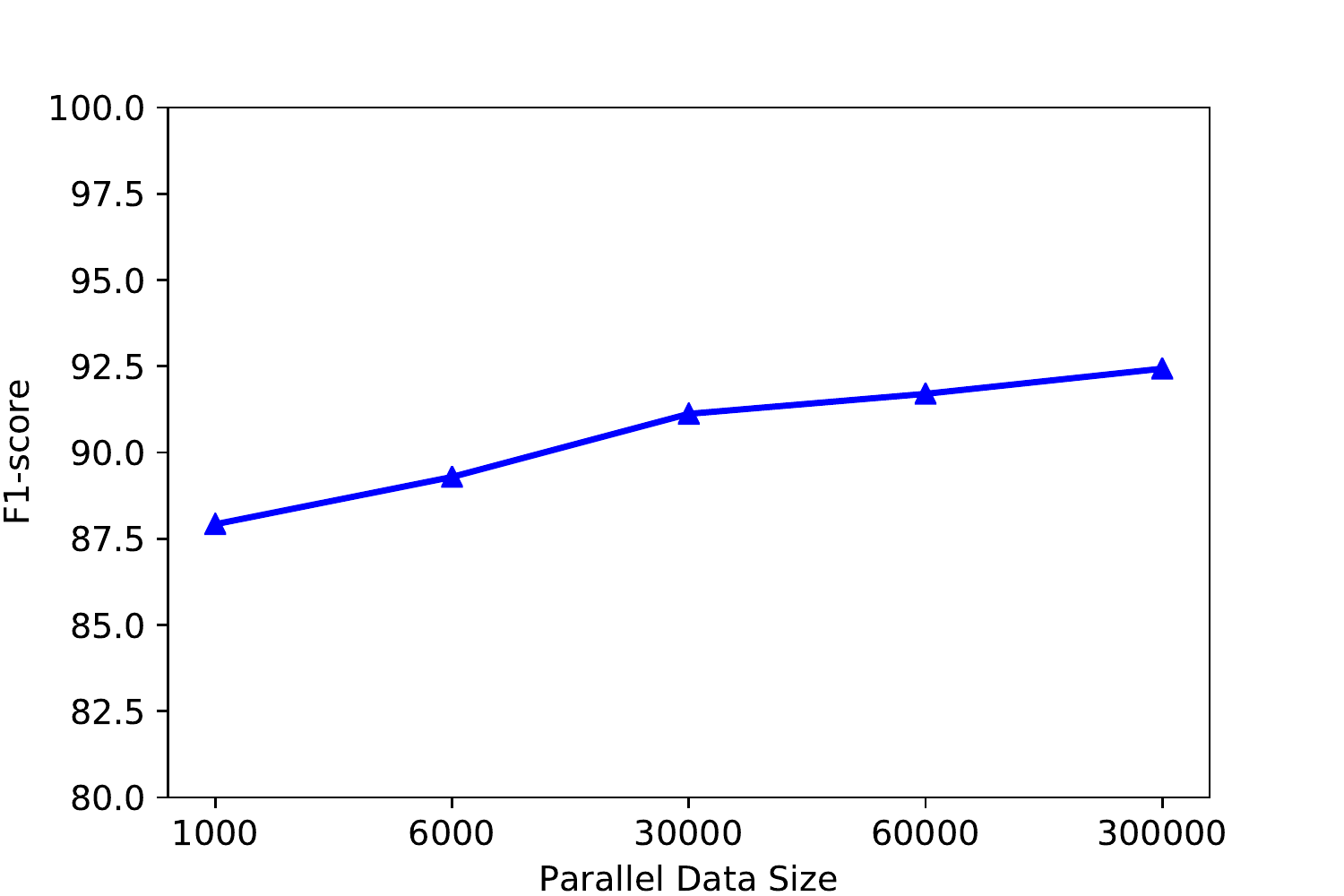}
	\caption{The F1 score of the key fact prediction model trained with different parallel data size.}\label{fig:table_curve}
\end{figure}

\begin{table}[t]
\centering
\begin{tabular}{l d{2.2} d{1.4} d{2.2}}
\toprule
\multicolumn{1}{c}{\bf Model} & \multicolumn{1}{c}{\bf BLEU} & \multicolumn{1}{c}{\bf NIST} & \multicolumn{1}{c}{\bf ROUGE} \\ 
\midrule
Vanilla Seq2Seq& 2.14 & 0.2809 & 0.47 \\
+ Pseudo & 10.01 & 3.0620 & 6.55 \\
\midrule
Transformer & 6.43 & 2.1019 & 1.77 \\
+ Pseudo & 14.35 &  4.1763 & 8.42 \\
\midrule
w/o Pseudo & 11.08 & 3.6910 & 4.84 \\
\Pivot-Vanilla & 20.09 &  6.5130 &  18.31  \\
\midrule
w/o Pseudo & 14.18 & 4.2686 & 7.10 \\
\Pivot-Trans & 27.34 & 6.8763 & 19.30\\
\bottomrule
\end{tabular}
\caption{Ablation study on the 1k training set for the effect of pseudo parallel data.}
\label{table:pseudo}
\end{table}

\begin{table}[t]
\centering
\begin{tabular}{l d{2.2} d{1.4} d{2.2}}
\toprule
\multicolumn{1}{c}{\bf Model} & \multicolumn{1}{c}{\bf BLEU} & \multicolumn{1}{c}{\bf NIST} & \multicolumn{1}{c}{\bf ROUGE} \\ 
\midrule
\Pivot-Vanilla & 20.09 &  6.5130 &  18.31 \\
w/o denosing & 18.45 & 4.8714 & 11.43 \\
\midrule
\Pivot-Trans & 27.34 & 6.8763 & 19.30 \\
w/o denosing & 25.72 &  6.5475 &  17.95 \\
\bottomrule
\end{tabular}
\caption{Ablation study on the 1k training set for the effect of the denoising data augmentation.}
\label{table:denosing}

\end{table}

\subsection{Varying Parallel Data Size}

We would like to further analyze the performance of our model given different size of parallel size. Therefore, we randomly shuffle the full parallel training set. Then, we extract the first $K$ samples as the parallel data, and modify the remaining data as the unlabeled data by removing the tables. We set $K=1000,6000,30000,60000,300000$, and compare our pivot models with both vanilla Seq2Seq and Transformer. Figure~\ref{fig:pivot_curve} shows the BLEU scores of our models and the baselines. When the parallel data size is small, the pivot model can outperform the vanilla Seq2Seq and Transformer by a large margin. With the increasement of the parallel data, the margin gets narrow because of the upper bound of the model capacity. Figure~\ref{fig:table_curve} shows the curve of the F1 score of the key fact prediction model trained with different parallel data size. Even when the number of annotated data is extremely small, the model can obtain a satisfying F1 score about $88\%$. In general, the F1 scores between the low and high parallel data sizes are close, which validates the assumption that the key fact prediction model does not rely on a heavy annotated data.

\subsection{Effect of Pseudo Parallel Data}

In order to analyze the effect of pseudo parallel data, we conduct ablation study by adding the data to the baseline models and removing them from our models. Table~\ref{table:pseudo} summarizes the results of the ablation study. 
Surprisingly, the pseudo parallel data can not only help the pivot model, but also significantly improve vanilla Seq2Seq and Transformer. The reason is that the pseudo parallel data can help the models to improve the ability of surface realization, which these models lack under the condition of limited parallel data. The pivot models can outperform the baselines with pseudo data, mainly because it breaks up the operation of key fact prediction and surface realization, both of which are explicitly and separately optimized.

\subsection{Effect of Denoising Data Augmentation}

We also want to know the effect of the denoising data augmentation. Therefore, we remove the denoising data augmentation from our model, and compare with the full model. Table~\ref{table:denosing} shows the results of the ablation study. It shows that the data augmentation brings a significant improvement to the pivot models under both vanilla Seq2Seq and Transformer frameworks, which demonstrates the efficiency of the denoising data augmentation.

\begin{table}[t]
\centering
    \begin{tabular}{p{7.2cm}}
    \hline
    \textbf{Transformer:}  
	a athletics -lrb- nfl -rrb- .  \\
    \hline
    \textbf{SemiMT:} 
	gustav dovid -lrb- born 25 august 1945 -rrb- is a former hungarian politician , who served as a member of the united states -lrb- senate -rrb- from president to 1989 . \\
    \hline
    \textbf{\Pivot-Trans:} 
	philippe adnot -lrb- born august 25 , 1945 -rrb- is a french senator , senator , and a senator of the french senate .\\
    \hline
    \hline
    \textbf{Reference:} 
    philippe adnot -lrb- born 25 august 1945 in rhèges -rrb- is a member of the senate of france . \\
    \hline 
    \end{tabular}
    \caption{An example of the generated text by our model and the baselines on 1k training set.}
    \label{tab:model_example}
\end{table}

\subsection{Qualitative Analysis}

We provide an example to illustrate the improvement of our model more intuitively, as shown in Table~\ref{tab:model_example}. Under the low resource setting, the Transformer can not produce a fluent sentence, and also fails to select the proper fact from the table. Thanks to the unlabeled data, the SemiMT model can generate a fluent, human-like description. However, it suffers from the hallucination problem so that it generates some unseen facts, which is not faithful to the source input. Although the \Pivot model has some problem in generating repeating words (such as ``senator'' in the example), it can select the correct key facts from the table, and produce a fluent description.

\section{Related Work}

This work is mostly related to both table-to-text generation and low resource natural language generation.

\subsection{Table-to-text Generation}

Table-to-text generation is widely applied in many domains. \citet{Dubou2002Content} proposed to generate the biography by matching the text with a knowledge base. \citet{Barzilay2005Collective} presented an efficient method for automatically learning content selection rules from a corpus and its related database in the sports domain. \citet{Liang2009Learning} introduced a system with a sequence of local decisions for the sportscasting and the weather forecast. Recently, thanks to the success of the neural network models, more work focused on the neural generative models in an end-to-end style~\citep{Wiseman2017Challenge,Puduppully2018D2T,Gehrmann2018E2E,Sha2018Order,Bao2018T2T,Qin2018Learning}. \citet{Lebret2016Neural} constructed a dataset of biographies from Wikipedia, and built a neural model based on the conditional neural language models. \citet{Liu2018T2T} introduced a structure-aware sequence-to-sequence architecture to model the inner structure of the tables and the interaction between the tables and the text. \citet{Wiseman2018Template} focused on the interpretable and controllable generation process, and proposed a neural model using a hidden semi-markov model decoder to address these issues. \citet{Nie2018Operation} attempted to improve the fidelity of neural table-to-text generation by utilizing pre-executed symbolic operations in a sequence-to-sequence model.

\subsection{Low Resource Natural Language Generation}

The topic of low resource learning is one of the recent spotlights in the area of natural language generation~\citep{Tilk2017Low,Tran2018Dual}. More work focused on the task of neural machine translation, whose models can generalize to other tasks in natural language generation. \citet{Gu2018Universal} proposed a novel universal machine translation which uses a transfer-learning approach to share lexical and sentence level representations across different languages. \citet{Cheng2016Semi} proposed a semi-supervised approach that jointly train the sequence-to-sequence model with an auto-encoder, which reconstruct the monolingual corpora. More recently, some work explored the unsupervised methods to totally remove the need of parallel data~\citep{Lample2018Phrase,Lample2017Unsupervised,Artetxe2017Unsupervised,zhang2018learning}.

\section{Conclusions}


In this work, we focus on the low resource table-to-text generation, where only limited parallel data is available. We separate the generation into two stages, each of which is performed by a model trainable with only a few annotated data. Besides, We propose a method to construct a pseudo parallel dataset for the surface realization model, without the need of any structured table. Experiments show that our proposed model can achieve $27.34$ BLEU score on a biography generation dataset with only $1,000$ parallel data.

\section*{Acknowledgement}

We thank the anonymous reviewers for their thoughtful comments. This work was supported in part by National Natural Science Foundation of China (No. 61673028). Xu Sun is the corresponding
author of this paper.

\bibliography{acl2019}
\bibliographystyle{acl_natbib}

\end{document}